\documentclass[10pt,twocolumn,letterpaper]{article}
\PassOptionsToPackage{table}{xcolor}
\usepackage{cvpr} % Show the author block in the CVPR layout.
\usepackage[T1]{fontenc}
\usepackage[utf8]{inputenc}
\usepackage{microtype}
\newcounter{medupalgorithm}
\usepackage{booktabs,multirow,makecell,pifont,enumitem,ragged2e}
\usepackage[skins,breakable]{tcolorbox}

\newcommand{\best}[1]{\textbf{#1}}
\newcommand{\second}[1]{\underline{#1}}
\definecolor{outerline}{HTML}{B5B5B5}
\definecolor{tokencolor}{HTML}{1F3A93}
\definecolor{boxblue}{HTML}{C8D2E0}
\definecolor{boxblueline}{HTML}{6A7DA0}
\definecolor{boxorange}{HTML}{F5D9B7}
\definecolor{boxorangeline}{HTML}{C58A3D}
\definecolor{boxgreen}{HTML}{CDE4C9}
\definecolor{boxgreenline}{HTML}{6FA864}
\definecolor{boxtitlebg}{HTML}{F5F7FA}
\definecolor{boxsubtitle}{HTML}{5B6B7A}
\definecolor{tableblue}{HTML}{EAF2FF}
\definecolor{tablegreen}{HTML}{EEF7EE}
\definecolor{tableorange}{HTML}{FFF4E8}
\definecolor{tablepurple}{HTML}{F4EEFF}
\definecolor{tablegray}{HTML}{F4F5F7}

\newtcolorbox{outerbox}{
  enhanced, colback=white, colframe=outerline,
  boxrule=0.5pt, arc=2mm,
  left=1.2mm, right=1.2mm, top=1.2mm, bottom=1.2mm,
  sidebyside, sidebyside align=top,
  lefthand width=0.30\linewidth,
  segmentation style={solid, draw=outerline, line width=0.4pt},
  bicolor, colbacklower=white,
}

\newtcolorbox{originalbox}{
  enhanced, colback=white, colframe=boxblueline,
  colbacktitle=boxblue, coltitle=black,
  fonttitle=\bfseries\footnotesize,
  title={Original Image},
  boxrule=0.4pt, arc=1mm,
  left=0.6mm, right=0.6mm, top=0.4mm, bottom=0.4mm,
  toptitle=0.3mm, bottomtitle=0.3mm,
}

\newtcolorbox{overlaybox}{
  enhanced, colback=white, colframe=boxorangeline,
  colbacktitle=boxorange, coltitle=black,
  fonttitle=\bfseries\footnotesize,
  title={Segmentation Overlay},
  boxrule=0.4pt, arc=1mm,
  left=0.6mm, right=0.6mm, top=0.4mm, bottom=0.4mm,
  toptitle=0.3mm, bottomtitle=0.3mm,
}

\newtcolorbox{cotbox}{
  enhanced, colback=white, colframe=boxgreenline,
  colbacktitle=boxgreen, coltitle=black,
  fonttitle=\bfseries\footnotesize,
  title={Chain-of-Thought Reasoning},
  boxrule=0.4pt, arc=1mm,
  left=1mm, right=1mm, top=0.6mm, bottom=0.6mm,
  toptitle=0.3mm, bottomtitle=0.3mm,
}

\definecolor{cvprblue}{rgb}{0.21,0.49,0.74}
\usepackage[pagebackref,breaklinks,colorlinks,allcolors=cvprblue]{hyperref}
\hypersetup{pdfauthor={Yuan Wang, Hualiang Wang, Yixin Chen, Songtao Jiang, Shujian Gao, Zhixiong Hu, Fucheng Niu, Xiaotian Zhang, Siming Fu, Jian Wu, Zuozhu Liu},pdftitle={MedUP: Bridging Medical Understanding and Perception with Adaptive Region Tokens}}

\definecolor{seghead}{HTML}{DEEAF6}
\definecolor{underhead}{HTML}{DFEFEA}
\definecolor{jointhead}{HTML}{EAE3F3}
\definecolor{costhead}{HTML}{F5E8D8}
\definecolor{ourswash}{HTML}{EDF4FA}
\definecolor{tableink}{HTML}{254B70}

\newcommand{\tablesetup}{\footnotesize\renewcommand{\arraystretch}{1.16}\setlength{\tabcolsep}{4pt}}

\newsavebox{\meduptablebox}
\NewDocumentEnvironment{spreadtable}{m m m +b}{%
  \sbox{\meduptablebox}{\begin{tabular}{#2}#4\end{tabular}}%
  \addtolength{\tabcolsep}{\dimexpr(#1-\wd\meduptablebox)/\numexpr#3+2\relax\relax}%
  \begin{tabular}{#2}#4\end{tabular}%
}{}
\def\confName{CVPR}
\def\confYear{2027}

\title{MedUP: Bridging Medical Understanding and Perception with \\ Adaptive Region Tokens}
\author{%
{Yuan Wang\textsuperscript{1,\textdaggerdbl}, Hualiang Wang\textsuperscript{1,\textdaggerdbl,*}, Yixin Chen\textsuperscript{1}, Songtao Jiang\textsuperscript{1}, Shujian Gao\textsuperscript{2}, Zhixiong Hu\textsuperscript{1},  }\\[3pt]
{Fucheng Niu\textsuperscript{1}, Xiaotian Zhang\textsuperscript{1}, Siming Fu\textsuperscript{1}, Jian Wu\textsuperscript{1}, Zuozhu Liu\textsuperscript{1,}\thanks{Corresponding authors. \quad \textsuperscript{\textdaggerdbl} Equal contributions.}}\\[5pt]
{\textsuperscript{1}Zhejiang University \quad \textsuperscript{2}Fudan University }
}

\usepackage{booktabs}
\usepackage{array}
\usepackage{multirow}
\DeclareMathSizes{7.2}{7.2}{5}{5}
\DeclareMathSizes{7.4}{7.4}{5}{5}
\definecolor{medjointhead}{HTML}{E8EFF6}
\definecolor{medourswash}{HTML}{F5ECDD}

\definecolor{invmod0}{HTML}{255A91}
\definecolor{invmod1}{HTML}{4A93C7}
\definecolor{invmod2}{HTML}{858D98}
\definecolor{invmod3}{HTML}{287F86}
\definecolor{invmod4}{HTML}{DF8B57}
\definecolor{invmod5}{HTML}{C69C38}
\definecolor{invmod6}{HTML}{AA854C}
\definecolor{invmod7}{HTML}{829C4D}
\definecolor{invmod8}{HTML}{4B8C71}
\definecolor{invmod9}{HTML}{459DA1}
\definecolor{invmod10}{HTML}{718B76}

\begin{document}
\maketitle

\begin{abstract}
Medical vision-language models (Med-VLMs) have advanced clinical question answering and report generation, yet their language understanding remains decoupled from spatial perception. Existing approaches represent locations as coordinates, invoke external segmentors, or decode masks through separate branches, none of which allow regions to be read, generated, and composed inside the language sequence itself. Discrete mask tokens resolve this by encoding masks into the language vocabulary, but existing tokenizers assign a fixed token budget to every mask: a short budget loses boundary detail on irregular lesions, whereas a long budget wastes sequence capacity on smooth organs. Treating these mask tokens as standalone outputs also leaves region perception uncoupled from clinical reasoning. We present \textbf{MedUP}, a medical VLM that bridges understanding and perception through \textbf{fidelity-constrained adaptive region tokens}. Its region tokenizer, \textbf{UniMedTok}, supervises cumulative residual-code prefixes and assigns each training mask the shortest prefix that preserves reconstruction fidelity. This adaptive interface allows the VLM to interleave text and variable-length region spans so that anatomical descriptions guide region perception and localized region evidence grounds clinical answers. Across our five-task \textbf{MedGround-Bench} evaluation, MedUP outperforms the strongest text-prompt baseline Citrus-V by 20.54 Dice points on 260K referring segmentation queries while using 58.5\% fewer residual codes per region than the full budget, and achieves 41.54 zero-shot Dice on the external MeCoVQA.
\end{abstract}

\section{Introduction}

\begin{figure}[t]
    \centering
    \includegraphics[width=0.95\columnwidth]{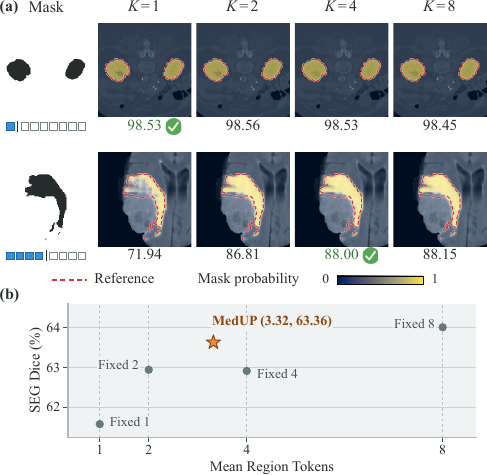}
    \vspace{-3mm}
    \caption{\textbf{Adaptive region-token lengths in MedUP.}
(a) Each row shows one reference mask reconstructed with increasing numbers of region codes.
The $\langle\mathrm{end}\rangle$~marker identifies the shortest training prefix whose reconstruction IoU is at least the eight-code IoU minus 0.01.
 (b) At inference on the full set, MedUP generates 3.32 codes per region on average and achieves 63.36 average Dice across 84 datasets, compared with 62.91 at a fixed budget of four codes. It remains only slightly below the eight-code fixed budget while using less than half as many codes.}
    \label{fig:region_token_budget}
\end{figure}

\label{sec:intro}
Fueled by large-scale medical vision-language corpora, medical vision-language models (Med-VLMs) have achieved strong performance in visual question answering, report generation, and clinical image interpretation~\citep{li2023llavamed,moor2023medflamingo,wu2023radfm,zhang2024biomedgpt,chen2024huatuogptvision}. Most of these systems follow a text-centered paradigm in which models verbalize what they see without delineating the underlying anatomy. Grounded medical interpretation requires language understanding and spatial perception at once. Clinical descriptions must refer to exact anatomical structures, and spatial predictions must follow the semantic target specified by users.

\begin{figure*}[t]
\centering
\includegraphics[width=0.95\textwidth]{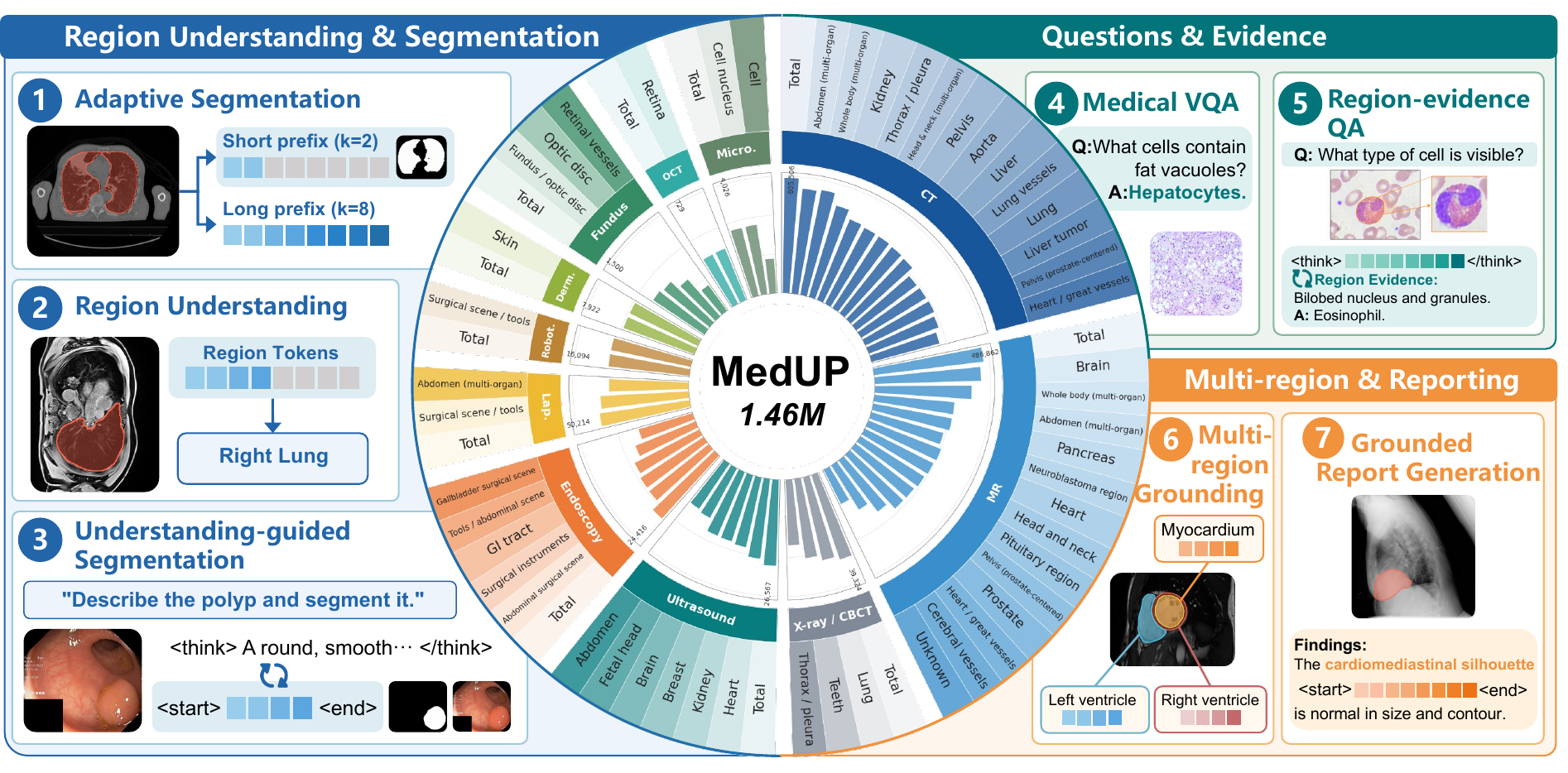}
\vspace{-3mm}
\caption{\textbf{Shared region tokens unify seven medical tasks across 11 modalities.} The seven training task views span adaptive segmentation, region understanding, understanding-guided segmentation, medical VQA, region-evidence QA, multi-region grounding, and grounded report generation. Training uses 1.46M image--mask pairs per epoch; a shared region-token interface connects anatomy and lesions with descriptions, answers, and reports.}
\label{fig:data_atlas}
\end{figure*}

Existing methods connect language to image regions through three main interfaces, each of which breaks sequence-level composition. Coordinate-based models express locations as bounding boxes or keypoints~\citep{chen2023shikra}, which indicate rough position but cannot capture the irregular boundary of a tumor or vessel. Tool-using agents delegate segmentation to external specialists~\citep{li2024mmedagent,jiang2026ibisagent} and leave the predicted mask outside the language model's autoregressive context. Grounded decoders attach a dedicated mask head to the language model~\citep{lai2024lisa,huang2025medplib,huang2025medsegr,luo-etal-2025-vividmed,wu2025unibiomed}, yet their dense mask outputs cannot be consumed as discrete tokens in subsequent reasoning steps. Discrete mask tokenization resolves this interface gap by encoding masks as vocabulary tokens for next-token prediction~\citep{wang2025himtok,zhou2026samtok}, but existing tokenizers such as SAMTok~\citep{zhou2026samtok} assign a fixed token budget to every mask. In Fig.~\ref{fig:region_token_budget}(a), a single token meets the reconstruction criterion for the smooth contours of the kidney pair, while the irregular necrotic tumor core needs four. A short fixed budget can lose detail along complex boundaries, while a long one spends tokens on simpler masks with little gain in fidelity. Treating these mask tokens as standalone outputs also leaves region perception uncoupled from clinical reasoning.

\begin{figure*}[t]
\centering
\IfFileExists{figures/pipeline_v3.pdf}{%
  \includegraphics[width=\textwidth]{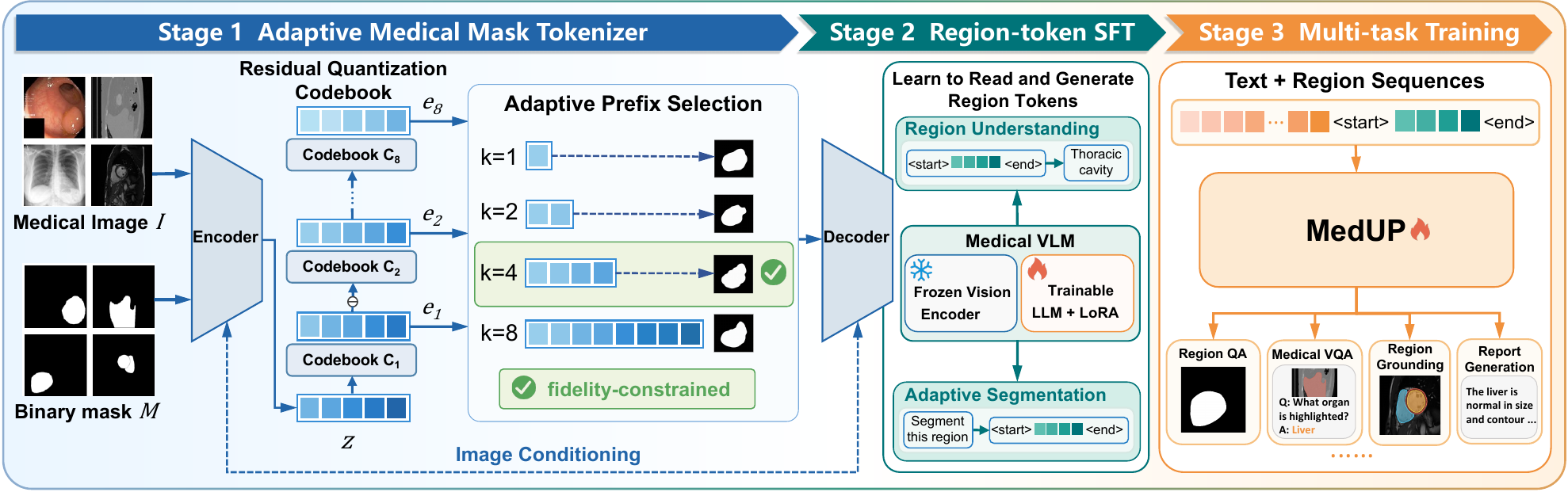}%
}{%
  \includegraphics[width=\textwidth]{figures/pipeline_v3.pdf}%
}
\vspace{-6mm}
\caption{\textbf{MedUP couples adaptive region tokenization with bidirectional region--language instruction tuning.} Stage 1 trains the tokenizer and selects the shortest fidelity-satisfying prefix.
Stage 2 aligns text-guided segmentation with region understanding; Stage 3 mixes five downstream views with the alignment views.
The VLM emits codes and termination at inference; the frozen decoder reconstructs masks.}\label{fig:pipeline}
\end{figure*}

Anatomical structures vary widely in boundary complexity, and regions encoded as variable-length language spans can both guide and be grounded by clinical reasoning. Based on this observation, We introduce \textbf{MedUP}, an effective medical VLM built on \textbf{UniMedTok}, an image-conditioned residual region tokenizer (Figs.~\ref{fig:data_atlas} and~\ref{fig:pipeline}). UniMedTok encodes each mask into ordered residual codes and supervises cumulative code prefixes during training. For each training mask, we formulate token allocation as a \emph{fidelity-constrained prefix selection} problem and select the shortest prefix whose reconstruction overlap stays within a tight tolerance of the full budget. During instruction tuning, the VLM learns to read and emit these variable-length region spans alongside text, and it predicts its own span termination at inference without oracle length hints. This shared autoregressive interface enables two complementary region--language supervision paths. \emph{Understanding-guided segmentation} generates an image-grounded anatomical description to distinguish the target from nearby anatomy before emitting its region span, while \emph{region-evidence question answering} localizes relevant image regions inside the reasoning chain before predicting the clinical answer.

Unlike grounded medical VLMs that decode masks through separate heads~\citep{huang2025medplib,wu2025unibiomed} and fixed-budget mask tokenizers that assign uniform code lengths~\citep{wang2025himtok,zhou2026samtok}, MedUP represents regions as variable-length language tokens whose budget adapts to anatomical complexity and supports multi-region composition in a single response. To evaluate these capabilities across single-region, multi-region, and compositional workflows, we establish \textbf{MedGround-Bench}, a unified benchmark covering referring segmentation and region understanding, medical VQA, multi-region grounding, region-evidence QA, and grounded report generation.

Trained on UniMedTok representations across 11 imaging categories, MedUP achieves 63.36 Dice on 260,280 referring segmentation queries in MedGround-Bench, outperforming the strongest text-prompt baseline Citrus-V by 20.54 Dice points and leading all compared models across every task (Tabs.~\ref{tab:main_v2} and~\ref{tab:segmentation_modality}). Our adaptive generation uses 58.5\% fewer residual codes per region than the full budget while preserving segmentation accuracy (Tab.~\ref{tab:adaptive}), and it achieves the highest zero-shot generalization on the external MeCoVQA benchmark with 41.54 modality-macro Dice (Tab.~\ref{tab:mecovqa_ood}). The training ablation confirms both paths of the mutual gain: target descriptions improve segmentation once region evidence is present, and region-span supervision improves QA once descriptions are present (Tab.~\ref{tab:factorial}). In summary, our contributions are:

\begin{itemize}[leftmargin=*,itemsep=2pt,topsep=3pt]
\item We introduce \textbf{UniMedTok} and \textbf{fidelity-constrained adaptive region tokens}, which select the matched prefix preserving reconstruction fidelity for efficient autoregressive generation and multi-region composition (Sec.~\ref{sec:method}).
\item We propose \textbf{bidirectional region--language supervision} through understanding-guided segmentation and region-evidence question answering, coupling anatomical perception with clinical reasoning (Sec.~\ref{sec:method_supervision}).
\item We establish \textbf{MedGround-Bench} for unified evaluation across five clinical tasks, where \textbf{MedUP} outperforms existing medical and grounded VLMs such as Citrus-V and UniPixel across all tracks (Sec.~\ref{sec:experiments}).
\end{itemize}

\section{Related Work}
\paragraph{Medical understanding and perception.}
Medical VLMs support question answering and interpretation~\citep{li2023llavamed,moor2023medflamingo,zhang2024biomedgpt,chen2024huatuogptvision,jiang2025hulumedtransparentgeneralistmodel,wang2025v2t,wang2026beyond}, while U-Net, nnU-Net, and medical adaptations of promptable segmentation models address region delineation~\citep{ronneberger2015unet,isensee2021nnunet,ma2024segment}. Grounded medical models connect language with spatial outputs~\citep{huang2025medplib,huang2025medsegr,luo-etal-2025-vividmed,wu2025unibiomed}. MedUP uses adaptive region tokens to test how targets affect segmentation and how region evidence affects understanding.

\vspace{-1em}
\paragraph{Region tokens and reasoning segmentation.}
LISA connects language reasoning to segmentation~\citep{lai2024lisa}, while HiMTok and SAMTok represent masks as discrete sequences for multimodal language models~\citep{wang2025himtok,zhou2026samtok}. MedUP uses decodable residual prefixes to adapt the region-token budget; its bidirectional region--language supervision further links anatomical descriptions to segmentation and localized visual evidence to clinical question answering.

\section{MedUP}
\label{sec:method}
% MedUP connects image regions with language through three components (Fig.~\ref{fig:pipeline}). Medical mask tokenization provides a discrete, decodable region interface. Adaptive-length representation allocates its capacity according to reconstruction fidelity. Instruction tuning then uses this interface to condition segmentation on descriptions and answers on region evidence.

MedUP links image regions and language through three components (Fig.~\ref{fig:pipeline}): a discrete, decodable mask-token interface; fidelity-adaptive region lengths; and instruction tuning that uses anatomical descriptions to guide segmentation and localized region evidence to support clinical answers.

\subsection{Medical Mask Tokenization}
UniMedTok is a MedSAM2-style image--mask residual tokenizer (MT256x8). For an image $I$ and region mask $M$, it produces ordered discrete residual codes $c_{1:8}$, with 256 entries per residual codebook. A prefix of length $k$ reconstructs a mask through an image-conditioned decoder:
\begin{equation}
 c_{1:8}=Q(E(I,M)),\qquad \hat M^{(k)}=D(I,c_{1:k}).
\end{equation}
The tokenizer is frozen for downstream instruction tuning. A region is serialized as a delimited sequence
\begin{equation}
 Z^{(k)}=[t_{\mathrm{s}},v_1(c_1),\ldots,v_k(c_k),t_{\mathrm{e}}],
\end{equation}
where $v_j$ maps a stage-specific code to its vocabulary token. $k$ counts residual mask codes; delimiter and text tokens are accounted for separately. Concatenating region spans and their associated text supports multiple targets without changing the underlying autoregressive interface. Image conditioning is essential: these codes serve as learned region prompts, rather than an image-independent lossless mask serialization.

\begin{table}[t]
\refstepcounter{medupalgorithm}
\label{alg:adaptive_prefix}
\small
\hrule\vspace{3pt}
\textbf{Algorithm \themedupalgorithm: Fidelity-Constrained Adaptive Prefix Selection}
\par\vspace{3pt}\hrule\vspace{4pt}
\textbf{Input:} Frozen tokenizer $(E,Q,D)$; regions $\mathcal D$; candidate lengths $\mathcal K$; tolerance $\epsilon\geq0$.\par
\textbf{Output:} Variable-length region targets $\{Z_i^*\}$.
\begin{enumerate}[label=\arabic*:,leftmargin=1.8em,itemsep=1pt,topsep=4pt,parsep=0pt]
\item $K_{\max}\gets\max\mathcal K$.
\item \textbf{for each} $(I_i,M_i)\in\mathcal D$ \textbf{do}
\item \hspace{0.5em}$c_{i,1:K_{\max}}\gets Q(E(I_i,M_i))$.
\item \hspace{0.5em}\textbf{for each} $k\in\mathcal K$ \textbf{do}
\item \hspace{1em}$\hat M_i^{(k)}\gets D(I_i,c_{i,1:k})$.
\item \hspace{1em}$q_i(k)\gets\operatorname{IoU}(\hat M_i^{(k)},M_i)$.
\item \hspace{0.5em}\textbf{end for}
\item \hspace{0.5em}$\mathcal F_i\gets\{k\in\mathcal K:q_i(k)\geq q_i(K_{\max})-\epsilon\}$.
\item \hspace{0.5em}$k_i^*\gets\min\mathcal F_i$.
\item \hspace{0.5em}$Z_i^*\gets[t_{\mathrm s},v_1(c_{i,1}),\ldots,v_{k_i^*}(c_{i,k_i^*}),t_{\mathrm e}]$.
\item \textbf{end for}
\item \textbf{return} $\{Z_i^*\}$.
\end{enumerate}
\hrule
\end{table}

\subsection{Adaptive-Length Region Representation}
\paragraph{Training decodable prefixes.}
Let $\mathcal K=\{1,2,4,8\}$. Multi-prefix tokenizer training decodes each of these cumulative prefixes for a training mask and averages their reconstruction losses, in addition to the quantization loss:
\begin{equation}
 \mathcal L_{\mathrm{tok}}=
 \frac{1}{|\mathcal K|}\sum_{k\in\mathcal K}
 \mathcal L_{\mathrm{rec}}(\hat M^{(k)},M)
 +\lambda_{\mathrm{vq}}\mathcal L_{\mathrm{vq}}.
\end{equation}
Each prefix incurs its own reconstruction loss, so the early codes receive supervision for recovering the mask even when later codes are absent. Tokenizer implementation details are provided in Supp.~\textcolor{cvprblue}{C}.
\begin{table*}[t]
  \centering
  \caption{\textbf{Unified evaluation on MedGround-Bench across five tasks.} Scores are percentages ($\uparrow$). Bold and underlining mark the highest and second-highest scores in each column, respectively. Region--language alignment covers referring segmentation and region understanding. Region understanding reports dataset-macro exact match (EM). Multi-region Dice weights targets equally, and AllC$_{.5}$ weights queries equally. Evidence QA and grounded report generation use dataset-macro averages. Joint$_{.5}$ requires a correct answer and evidence IoU $\geq0.5$. A dash denotes unavailable output capability. Dataset-level results and protocols appear in Supp.~\textcolor{cvprblue}{G}.}
    \label{tab:main_v2}
  \tablesetup
  \begin{spreadtable}{\dimexpr0.95\textwidth-2pt\relax}{lccccccccccc}{22}
  \toprule
  \multirow{2}{*}{Method} & \multirow{2}{*}{Size} &
  \multicolumn{2}{c}{\cellcolor{seghead}\textbf{Referring SEG}} &
  \cellcolor{underhead}\textbf{Region U.} &
  \cellcolor{underhead}\textbf{Med. VQA} &
  \multicolumn{2}{c}{\cellcolor{jointhead}\textbf{Multi-region}} &
  \multicolumn{2}{c}{\cellcolor{jointhead}\textbf{Evidence QA}} &
  \multicolumn{2}{c}{\cellcolor{jointhead}\shortstack{\textbf{Grounded Report Gen.}}}\\
  \cmidrule(lr){3-4}
  \cmidrule(lr){5-5}
  \cmidrule(lr){6-6}
  \cmidrule(lr){7-8}
  \cmidrule(lr){9-10}
  \cmidrule(l){11-12}
  & & IoU & Dice & EM & Score &
  Dice & AllC$_{.5}$ & Acc. & Joint$_{.5}$ &
  ROUGE-L & Region Dice\\

  \midrule
  \multicolumn{12}{@{}l}{\textit{Medical and general vision--language models}}\\[1pt]

  Qwen2.5-VL~\cite{bai2025qwen25vltechnicalreport} & 3B &
  - & - & 28.65 & 35.34 &
  - & - & 50.72 & - & 8.13 & -\\

  Qwen3-VL~\cite{bai2025qwen3} & 4B &
  - & - & 40.26 & 35.49 &
  - & - & \second{65.81} & - & 9.40 & -\\

  LLaVA-Med~\cite{li2023llavamed} & 7B &
  - & - & 30.09 & 28.80 &
  - & - & 43.03 & - & 8.59 & -\\

  HuatuoGPT-Vision~\cite{chen2024huatuogptvision} & 7B &
  - & - & 36.77 & 35.45 &
  - & - & 58.34 & - & 8.16 & -\\

  \addlinespace[3pt]
  \multicolumn{12}{@{}l}{\textit{Grounded and pixel-language models}}\\[1pt]

  BioMedParse~\cite{zhao2025foundation} & 0.4B &
  11.02 & 14.16 & - & - &
  21.08 & 2.50 & - & - & - & 1.39\\

  Citrus-V~\cite{wang2025citrusv} & 8B &
  \second{36.56} & \second{42.82} & \second{42.39} & \second{55.65} &
  1.48 & 0.00 & 31.18 & 0.03 & 11.36 & 4.36\\

  UniBiomed~\cite{wu2025unibiomed} & 2B &
  24.94 & 31.20 & 18.70 & 7.70 &
  1.20 & 0.00 & 1.18 & 0.01 & 8.91 & 0.49\\

  LISA++~\cite{yang2023lisa++} & 7B &
  9.20 & 13.72 & 0.00 & 30.51 &
  15.75 & 0.22 & 17.84 & 0.09 & 4.88 & 0.60\\

  SAM4MLLM~\cite{chen2024sam4mllm} & 8B &
  6.13 & 9.87 & 11.15 & 21.58 &
  8.39 & 0.00 & 14.27 & 0.00 & 9.48 & 3.13\\

  MMedAgent~\cite{li2024mmedagent} & 7B &
  20.25 & 25.10 & 12.42 & 7.06 &
  - & - & 15.75 & - & 3.74 & -\\

  MedPLIB~\cite{huang2025medplib} & 7B &
  31.22 & 38.73 & 38.36 & 54.65 &
  0.03 & 0.00 & 58.95 & 2.00 & 6.23 & 0.48\\

  PixelLM~\cite{ren2024pixellm} & 7B &
  9.01 & 13.67 & 23.63 & 11.91 &
  13.39 & 0.33 & 16.49 & 0.00 & 5.98 & 2.14\\

  UniPixel~\cite{liu2026unipixel} & 7B &
  17.06 & 23.00 & 32.51 & 28.18 &
  \second{24.65} & 4.57 & 34.53 & 1.67 & \second{12.87} & 3.52\\

  VisionReasoner~\cite{liu2025visionreasoner} & 7B &
  15.98 & 21.46 & 37.44 & 37.45 &
  20.55 & \second{7.26} & 55.17 & \second{2.59} & 10.19 & \second{10.29}\\

  \midrule

  \rowcolor{ourswash}

  MedUP (Ours) & 4B &
  \best{55.68} & \best{63.36} & \best{55.09} & \best{64.75} &
  \best{75.31} & \best{50.98} & \best{70.89} & \best{7.55} & \best{39.23} & \best{13.99}\\

  \bottomrule
  \end{spreadtable}

  \end{table*}

\paragraph{Fidelity-constrained prefix selection.}
We minimize the number of residual codes needed to preserve reconstruction fidelity. For region $i$, let $q_i(k)=\operatorname{IoU}(D(I_i,c_{i,1:k}),M_i)$ and $K_{\max}=\max\mathcal K=8$. The target length solves
\begin{equation}
\begin{aligned}
 k_i^*&=\underset{k\in\mathcal K}{\arg\min}\;k\\
 \text{s.t.}\quad q_i(k)&\geq q_i(K_{\max})-\epsilon,
\end{aligned}
\label{eq:kstar}
\end{equation}
Here $\epsilon=0.01$ is an absolute IoU tolerance on $[0,1]$. The eight-code prefix always meets the constraint. Because IoU need not increase at every step, Algorithm~\ref{alg:adaptive_prefix} scores all four prefixes and selects the shortest feasible one for each mask. The constraint applies to each region independently, so the sum of the selected lengths is minimal under these per-region tolerances. Absolute reconstruction-quality filtering is applied separately (Supp.~\textcolor{cvprblue}{C}).

\vspace{-1em}
\paragraph{Generated length at inference.}
The oracle target $k_i^*$ uses ground-truth masks only during offline target construction. The VLM learns region-token content and the end delimiter from these sequences; in adaptive-auto inference, generated termination determines the emitted span length without access to the target mask or its oracle length. We distinguish \emph{Oracle-$K^*$}, \emph{Adaptive-auto}, and fixed-budget generation throughout evaluation. Implementation and decoding details are given in Supp.~\textcolor{cvprblue}{C}.

\subsection{Region--Language Supervision}
\label{sec:method_supervision}
Because UniMedTok serializes masks as vocabulary tokens, MedUP models both text and image regions within a single autoregressive sequence $Y$ conditioned on image $I$ and multimodal prompt $X$:
\begin{equation}
 p_\theta(Y\mid I,X)=\prod_{t=1}^{|Y|}p_\theta(y_t\mid I,X,y_{<t}).
\end{equation}
For region understanding, $X$ embeds an input region span $Z^{(k)}$; for region generation and grounded reasoning, variable-length region spans $Z^{(k^*)}$ are emitted directly inside $Y$. This unified sequence interface enables two complementary supervision views on paired examples sharing the same image and anatomical target (Supp.~\textcolor{cvprblue}{B}).

\paragraph{Understanding-guided segmentation ($\mathrm{U}\to\mathrm{S}$).}
Directly emitting mask tokens from a short clinical name often fails when the target shares low-contrast boundaries with adjacent structures. To ground perception in semantic context, the understanding-guided view supervises $Y=[R;Z^{(k^*)}]$, where the model first generates an image-grounded anatomical description $R$ specifying appearance and location before emitting the adaptive region span $Z^{(k^*)}$. Its paired direct view supervises $Y=Z^{(k^*)}$ on the identical image, prompt, and target mask.
\vspace{-1mm}
\paragraph{Region-evidence question answering ($\mathrm{S}\to\mathrm{U}$).}
Conversely, clinical question answering is prone to ungrounded guessing when predictions are not anchored to the relevant lesion or organ. The region-evidence view supervises $Y=[E_{\mathrm{reg}};A]$, where $E_{\mathrm{reg}}$ is an analytical reasoning chain that interleaves one or more predicted evidence region spans $Z^{(k^*)}$ prior to the final answer $A$. For the paired text-only view (Supp.~\textcolor{cvprblue}{B}), we omit the region spans but retain the same image, question, reasoning text, and answer $A$.

\paragraph{Multi-task supervision.}
Across Stages 2 and 3, instruction supervision covers seven task views. Stage 2 focuses on text-guided segmentation and region understanding. Stage 3 combines examples from both tasks with supervision for understanding-guided segmentation, medical VQA, multi-region grounding, region-evidence QA, and grounded report generation. For the selected records $\mathcal D$, the supervised objective is
\begin{equation}
 \mathcal L_{\mathrm{SFT}}=-\mathbb E_{(I,X,Y)\sim\mathcal D}
 \sum_{t\in\mathcal A}\log p_\theta(y_t\mid I,X,y_{<t}),
\end{equation}
where $\mathcal A$ indexes supervised assistant positions. All three stages use supervised objectives. See \mbox{Supp.~\textcolor{cvprblue}{A} and~\textcolor{cvprblue}{C}} for stage-wise data and implementation details.
\section{Experiments}
\label{sec:experiments}

\vspace{-1mm}
\subsection{Experimental Setup}
\label{sec:exp_setup}

\paragraph{Training Data and Benchmarks.}
MedUP uses Qwen3-VL-4B as its vision-language backbone across three supervised stages. Stage 1 trains the UniMedTok residual tokenizer on 1.46M image-mask pairs per epoch from 123 dataset sources across 11 imaging modalities for 10 epochs, jointly supervising cumulative prefixes $K\in\{1,2,4,8\}$. With the tokenizer frozen, Stage 2 aligns the VLM with the region vocabulary for one epoch on 2.10M instruction pairs covering text-guided segmentation and region understanding, and Stage 3 fine-tunes the aligned model for one epoch on approximately 410K instruction pairs, combining five downstream tasks with segmentation and region-understanding examples from the alignment stage. We use the original dataset splits. For in-domain evaluation, we establish \textbf{MedGround-Bench}, which organizes evaluation into five core region-language tasks: \textbf{region--language alignment} covering referring segmentation and region understanding across eight imaging modality groups, \textbf{medical VQA} on SLAKE, PathVQA, and VQA-RAD, \textbf{multi-region grounding} across CT, MRI, and surgical-video benchmarks, \textbf{evidence QA} on ClinGroundQA, VinDr~\citep{nguyen2020vindr,nguyen2025heal}, and OmniMedVQA~\citep{hu2024omnimedvqa}, and \textbf{grounded report generation} on MS-CXR and IU-Xray. For external evaluation, we test zero-shot segmentation across five modalities on \textbf{MeCoVQA}~\citep{huang2025medplib}. We checked image MD5 hashes and reviewed the provenance of all public data sources to exclude identifiable case overlap between training and either benchmark, and between the two benchmarks (Supp.~\textcolor{cvprblue}{A}). Exact sample counts, source inventories, and dataset curation protocols for MedGround-Bench and MeCoVQA are detailed in Supp.~\textcolor{cvprblue}{A} and~\textcolor{cvprblue}{D}--\textcolor{cvprblue}{F}.
\vspace{-1em}
\paragraph{Baselines.}
Tab.~\ref{tab:main_v2} compares general and medical VLMs on text-output tasks and grounded models on tasks supported by their outputs. Text-prompted models receive only the image and instruction. MedSAM, MedSAM2, and MedSAM3 receive ground-truth boxes expanded by $1.25\times$ and are reported separately (Tab.~\ref{tab:segmentation_modality}). See Supp.~\textcolor{cvprblue}{G} for the adapters.
%%%%%%%%%%%%%%%
% 要不要提1.25x？
%%%%%%%%%%%%%%%

\begin{table*}[t]
\centering
\caption{\textbf{Medical segmentation across imaging modalities.} Evaluation uses 260,280 category-union queries from 84 datasets. Overall Dice, IoU, NSD, and modality-specific Dice are per-query means (\%, $\uparrow$); NSD uses a five-pixel tolerance. GT-box and text-prompt methods are compared separately. US: ultrasound; GI: gastrointestinal endoscopy; Surg.: surgical endoscopy; Derm.: dermoscopy. Bold and underlining indicate the best and second-best listed scores within each prompt group.}
\label{tab:segmentation_modality}
\tablesetup
\begin{spreadtable}{\dimexpr0.95\textwidth-5pt\relax}{lccccccccccc}{22}
\toprule
\multirow{2}{*}{Method} & \multicolumn{3}{c}{\cellcolor{seghead}\textbf{Overall}} & \multicolumn{8}{c}{\cellcolor{underhead}\textbf{Dice by modality}}\\
\cmidrule(lr){2-4}\cmidrule(l){5-12}
& Dice & IoU & NSD & CT & MRI & US & GI & Surg. & Fundus & Derm. & X-ray\\
\midrule
\multicolumn{12}{@{}l}{\textit{GT-box prompts}}\\[1pt]
MedSAM~\cite{ma2024segment} & \textbf{70.07} & \textbf{59.30} & \textbf{71.39} & \textbf{74.14} & \textbf{62.14} & \textbf{85.63} & 89.79 & \textbf{70.81} & \textbf{87.13} & \underline{86.56} & 67.47\\
MedSAM2~\cite{ma2025medsam2} & \underline{62.78} & \underline{51.97} & \underline{66.19} & \underline{62.48} & \underline{59.33} & 57.95 & \underline{90.17} & \underline{69.91} & \underline{83.54} & 82.36 & \textbf{74.59}\\
MedSAM3~\cite{liu2025medsam3} & 51.38 & 42.18 & 55.25 & 51.50 & 48.45 & \underline{67.95} & \textbf{91.10} & 52.74 & 67.01 & \textbf{90.08} & \underline{68.66}\\
\addlinespace[3pt]
\multicolumn{12}{@{}l}{\textit{Text prompts: baselines}}\\[1pt]
BioMedParse~\cite{zhao2025foundation} & 14.16 & 11.02 & 17.74 & 17.70 & 15.17 & 25.14 & 0.26 & 0.14 & 0.00 & 0.01 & 0.25\\
Citrus-V~\cite{wang2025citrusv} & \underline{42.82} & \underline{36.56} & \underline{46.17} & \underline{56.50} & \underline{38.63} & 17.22 & \textbf{92.44} & 4.08 & \underline{58.89} & \textbf{89.58} & \underline{60.23}\\
UniBiomed~\cite{wu2025unibiomed} & 31.20 & 24.94 & 33.27 & 34.82 & 35.38 & \underline{43.79} & 75.91 & 5.19 & 24.27 & 72.86 & 53.64\\
LISA++~\cite{yang2023lisa++} & 13.72 & 9.20 & 7.53 & 11.35 & 9.45 & 30.42 & 60.31 & 26.34 & 24.71 & 69.40 & 15.81\\
SAM4MLLM~\cite{chen2024sam4mllm} & 9.87 & 6.13 & 2.60 & 8.82 & 7.23 & 23.13 & 42.81 & 14.57 & 25.65 & 63.53 & 19.84\\
MMedAgent~\cite{li2024mmedagent} & 25.10 & 20.25 & 23.98 & 30.88 & 20.45 & 17.66 & 50.99 & 13.58 & 43.98 & 67.37 & 26.79\\
MedPLIB~\cite{huang2025medplib} & 38.73 & 31.22 & 36.83 & 50.68 & 34.10 & 42.86 & 27.10 & 7.48 & 40.70 & 79.12 & 26.59\\
PixelLM~\cite{ren2024pixellm} & 13.67 & 9.01 & 5.47 & 10.51 & 8.63 & 33.31 & 27.41 & 31.60 & 27.46 & 50.17 & 24.01\\
UniPixel~\cite{liu2026unipixel} & 23.00 & 17.06 & 15.90 & 17.92 & 20.36 & 24.76 & 64.83 & \underline{42.08} & 50.12 & 77.16 & 36.21\\
VisionReasoner~\cite{liu2025visionreasoner} & 21.46 & 15.98 & 15.48 & 18.61 & 15.12 & 32.93 & 71.73 & 40.14 & 51.56 & 78.90 & 39.08\\
\midrule
\rowcolor{ourswash}
MedUP (Ours) & \textbf{63.36} & \textbf{55.68} & \textbf{64.70} & \textbf{70.71} & \textbf{49.29} & \textbf{76.54} & \underline{87.48} & \textbf{66.69} & \textbf{79.36} & \underline{87.57} & \textbf{60.44}\\
\bottomrule
\end{spreadtable}
\end{table*}

\paragraph{Metrics.}
All quality scores are reported as percentages. Referring segmentation uses per-query mean Dice, IoU, and five-pixel Normalized Surface Dice, while MeCoVQA uses unweighted macro-average Dice across its five modalities. For region understanding, we compute exact match (EM) within each dataset and report the unweighted mean across datasets for every model. Medical VQA pools closed-ended accuracy and open-ended exact match. Multi-region grounding reports per-region Dice and query-level all-regions-correct rate AllC$_{.5}$, which requires matching the target region count and reaching IoU $\geq 0.5$ on every target. Evidence QA averages VinDr, OmniMedVQA, and ClinGroundQA equally on answer accuracy, evidence Dice, and joint success Joint$_{.5}$, which requires both a correct answer and evidence IoU $\geq 0.5$. Grounded report generation averages MS-CXR and IU-Xray equally on ROUGE-L and region Dice, with BLEU and METEOR reported in Supp.~Tab.~\textcolor{cvprblue}{S10}. For adaptive tokenization, we report mean residual codes per region $\bar K$, reference-mask reconstruction Dice, fidelity-violation rate $V$ at tolerance $\epsilon=0.01$ relative to $K=8$, and downstream segmentation Dice.

\begin{figure*}[t]
\centering
\includegraphics[width=\textwidth]{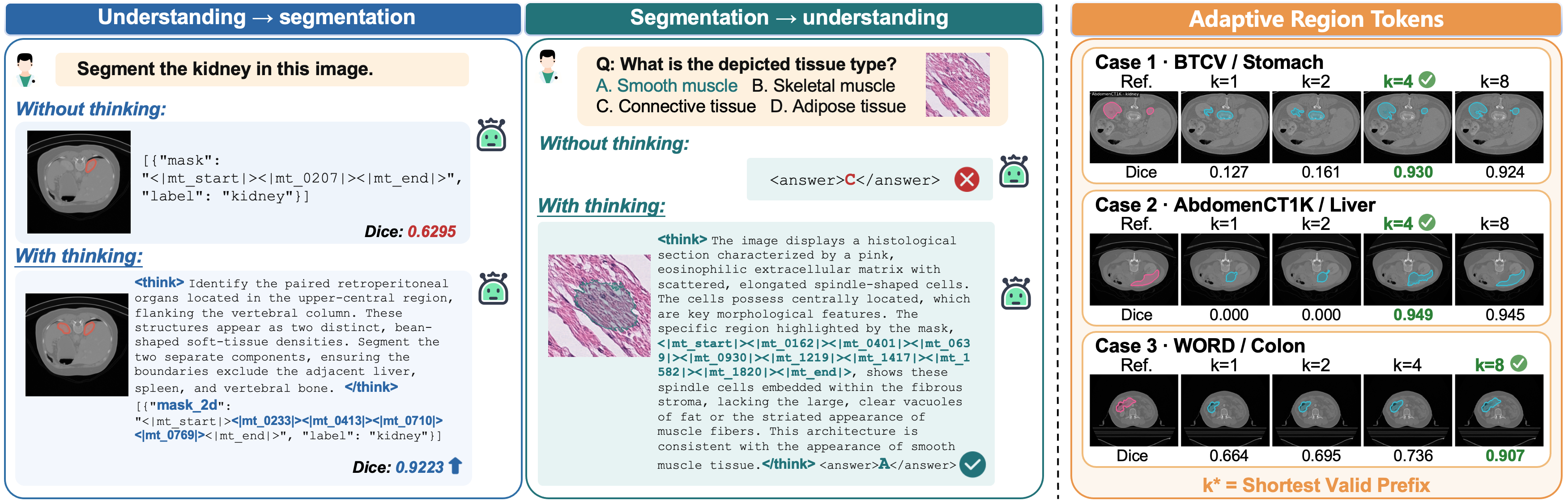}
\vspace{-7mm}
\caption{\textbf{Adaptive prefix selection, understanding-guided segmentation, and region-evidence QA.} Left, adaptive prefix selection: check marks give the shortest fidelity-satisfying prefix (Eq.~\ref{eq:kstar}), with Dice below each reconstruction.
Center, describing the target first raises kidney Dice from 0.6295 to 0.9223.
Right, region evidence corrects the answer from connective tissue to the reference label, smooth muscle.}
\label{fig:thinking_cases}
\end{figure*}

\vspace{-1mm}

\subsection{Main Results}
\label{sec:exp_main}
\vspace{-1mm}
\paragraph{Referring segmentation across modalities.}
On MedGround-Bench (Tabs.~\ref{tab:main_v2} and~\ref{tab:segmentation_modality}), MedUP reaches 63.36 Dice, 55.68 IoU, and 64.70 NSD with a 4B backbone, outperforming the strongest text-prompt baseline Citrus-V-8B by 20.54 Dice, 19.12 IoU, and 18.53 NSD points. Without spatial prompts at inference, MedUP also surpasses box-prompted MedSAM2 at 62.78 Dice and MedSAM3 at 51.38 Dice, approaching GT-box MedSAM at 70.07 Dice. Across the eight modality groups in Tab.~\ref{tab:segmentation_modality}, MedUP ranks first among text-prompted models in CT, MRI, ultrasound, surgical endoscopy, fundus, and X-ray, gaining $+32.75$ Dice over UniBiomed on ultrasound and $+24.61$ Dice over UniPixel on surgical endoscopy, while ranking second to Citrus-V on gastrointestinal endoscopy and dermoscopy. Qualitative comparisons are provided in Supp.~Fig.~\textcolor{cvprblue}{S1}.

% Evaluation covers 2,791 regions for reference reconstruction and 4,056 queries for downstream segmentation.

% At test time, each VLM predicts its own span length.
\begin{table*}[htbp]
\centering
\caption{\textbf{Offline allocation of tokenizer prefixes to reference masks.} Target $\bar K$ is the mean number of residual codes assigned per reference region. A violation occurs when reconstruction IoU falls more than 0.01 below the eight-code reconstruction IoU for that region. The final column gives segmentation Dice for a VLM trained on each policy's targets. See Supp.~\textcolor{cvprblue}{C.1} for the evaluation protocol.}
\label{tab:allocation_policy}
\small
\setlength{\tabcolsep}{5pt}
\renewcommand{\arraystretch}{1}
\begin{tabular*}{0.8\textwidth}{@{\extracolsep{\fill}}lcccc@{}}
\toprule
Offline target policy
& Target $\bar K\downarrow$
& Recon. Dice $\uparrow$
& Violation (\%) $\downarrow$
& SEG Dice $\uparrow$\\
\midrule
Fixed $K=2$
& 2.000 & 86.816 & 64.959 & 34.003\\
Fixed $K=4$
& 4.000 & 89.813 & 43.318 & 31.231\\
Fixed $K=8$
& 8.000 & 90.828 & \textbf{0.000} & 30.260\\
Random allocation (training-reference sampling)
& 4.696 & 87.325 & 37.241 & 29.039\\
Shuffled adaptive lengths (matched histogram)
& 4.612 & 87.430 & 37.764 & 30.482\\
\rowcolor{ourswash}
\textbf{Fidelity-constrained selection}
& 4.612 & \textbf{91.192} & \textbf{0.000} & \textbf{40.304}\\
\bottomrule
\end{tabular*}
\end{table*}
\vspace{-1em}
\paragraph{Region understanding and medical VQA.}
When interpreting input region tokens in Tab.~\ref{tab:main_v2}, MedUP achieves 55.09 dataset-macro EM on region understanding, exceeding Citrus-V by 12.70 points and its base backbone Qwen3-VL-4B by 14.83 points, with per-modality results in Supp.~Tab.~\textcolor{cvprblue}{S1}. On medical VQA, MedUP reaches 64.75 and outperforms Citrus-V by 9.10 points and MedPLIB by 10.10 points, confirming that region-language alignment strengthens visual question answering rather than trading off text capability for segmentation.

\vspace{-2mm}
\paragraph{Multi-region grounding.}
MedUP reaches 75.31 region Dice and 50.98\% AllC$_{.5}$ (Tab.~\ref{tab:main_v2}), exceeding UniPixel by 50.66 Dice points and VisionReasoner by 43.72 AllC$_{.5}$ points. AllC$_{.5}$ requires the correct number of masks and an IoU of at least 0.5 for every named target. These scores evaluate mask localization and target binding. Spatial relation text is unscored.
\vspace{-2mm}
\paragraph{Evidence QA.}
Averaged equally across ClinGroundQA, VinDr, and OmniMedVQA (Tab.~\ref{tab:main_v2}), MedUP achieves 70.89\% answer accuracy and 7.55\% Joint$_{.5}$, surpassing its base backbone Qwen3-VL-4B by 5.08 accuracy points and nearly tripling the strongest grounded baseline VisionReasoner in Joint$_{.5}$ (7.55\% vs.\ 2.59\%; Citrus-V reaches 0.03\%). Joint$_{.5}$ is a strict composite metric requiring both a correct clinical answer and an unprompted evidence mask with IoU $\geq 0.5$ against reference annotations (physician freehand masks on ClinGroundQA and box-derived masks on VinDr and OmniMedVQA; Supp.~Tab.~\textcolor{cvprblue}{S12}). Alongside box-derived VinDr and OmniMedVQA for multi-modality coverage, we evaluate separately on physician-annotated ClinGroundQA: MedUP answers 311 of 383 questions correctly (81.20\%) and achieves the highest Joint$_{.5}$ (7.05\% vs.\ 6.01\% for MedPLIB and 4.70\% for VisionReasoner), whereas existing grounded VLMs rarely localize evidence without explicit segmentation prompts.
\vspace{-3mm}
\paragraph{Grounded report generation.}
MedUP reaches 39.23 ROUGE-L and 13.99 region Dice, outperforming UniPixel by 26.36 ROUGE-L points and VisionReasoner by 3.70 Dice points. Per-dataset results for the composed tasks appear in Supp.~Tabs.~\textcolor{cvprblue}{S10}--\textcolor{cvprblue}{S12}.

\subsection{Adaptive Region Tokenization}
\label{sec:exp_adaptive}

\paragraph{Reference-mask reconstruction.}
To separate mask representation quality from VLM prediction errors, we first evaluate reference-mask reconstruction with the frozen UniMedTok decoder in Tab.~\ref{tab:adaptive}. Mean reconstruction Dice rises from 88.44 at $K=1$ to 92.04 at $K=8$.

\vspace{-3mm}
% 原则上可以easy模式少token，复杂的过程需要更长的token
\paragraph{Adaptive generation vs.\ fixed token budgets.}
In Tab.~\ref{tab:adaptive}, Adaptive-auto generates 3.32 residual codes per region on average, 58.5\% fewer than the eight-code budget. Its segmentation Dice is 63.36, 0.65 points below the 64.01 obtained at $K=8$. Adaptive-auto has the highest MedVQA accuracy (64.75\%), 0.27 percentage points above the best fixed budget ($K=2$). Decoding reference codes at the predicted lengths gives 92.34 reconstruction Dice. This reconstruction score measures the fidelity available at the chosen lengths. Downstream Dice also depends on the generated codes.
\vspace{-1em}

\begin{table}[t]
\centering
\vspace{-2mm}
\caption{\textbf{VLM generation at fixed and adaptive region budgets.} The code count excludes delimiters. Fixed rows give the specified budget, and Adaptive-auto reports the mean length generated by the VLM. Reconstruction Dice uses a reference-mask prefix of the indicated length. SEG Dice uses codes predicted from the image and instruction. QA Acc. measures MedVQA answer accuracy.}
\label{tab:adaptive}
\tablesetup\setlength{\tabcolsep}{3pt}
  \begin{spreadtable}{0.95\columnwidth}{lcccc}{8}
  \toprule
  \multirow{2}{*}{Budget}
  & \cellcolor{costhead}\textbf{Codes}
  & \cellcolor{seghead}\textbf{Recon.}
  & \multicolumn{2}{c}{\cellcolor{jointhead}\textbf{Downstream}}\\
  \cmidrule(lr){2-2}\cmidrule(lr){3-3}\cmidrule(l){4-5}
  & $\bar K\!\downarrow$
  & Dice$\uparrow$
  & SEG Dice$\uparrow$
  & QA Acc.$\uparrow$\\
  \midrule
  Fixed $K=1$ & 1 & 88.44 & 61.57 & 62.34\\
  Fixed $K=2$ & 2 & 91.03 & 62.94 & 64.48\\
  Fixed $K=4$ & 4 & 91.51 & 62.91 & 63.46\\
  Fixed $K=8$ & 8 & 92.04 & \textbf{64.01} & 63.32\\
  \midrule
  \rowcolor{ourswash}
  \textbf{Adaptive-auto} & 3.32 & \textbf{92.34} & 63.36 & \textbf{64.75}\\
  \bottomrule
  \end{spreadtable}
\vspace{3pt}
\end{table}

\begin{table}[htbp]
\centering
\vspace{-2mm}
\caption{\textbf{External segmentation generalization on MeCoVQA.} Dice (\%, $\uparrow$) is evaluated on 1,243 queries across five modalities. Avg. is the unweighted mean of modality-level scores. US: ultrasound; Endo.: endoscopy. Bold and underlining indicate the best and second-best scores, respectively.}
\label{tab:mecovqa_ood}
\begingroup
\setlength{\tabcolsep}{2pt}
\renewcommand{\arraystretch}{1.16}
\scriptsize
\begin{tabular*}{0.95\columnwidth}{@{\extracolsep{\fill}}lrrrrrr@{}}
\toprule
& \multicolumn{5}{c}{\cellcolor{seghead}\textbf{Dice by modality}} & \cellcolor{underhead}\textbf{Avg.}\\
\cmidrule(lr){2-6}\cmidrule(l){7-7}
Method & MRI & X-ray & US & Fundus & Endo. & Dice\\
\midrule
MedPLIB & \textbf{28.02} & 6.64 & \textbf{33.94} & 1.57 & 42.86 & 22.61\\
BioMedParse & 3.05 & 0.51 & 0.00 & 0.00 & 0.21 & 0.75\\
UniBiomed & 22.50 & \textbf{30.18} & 0.20 & \underline{67.86} & \underline{66.09} & \underline{37.37}\\
LISA++ & 10.23 & 3.80 & 6.08 & 3.36 & 47.93 & 14.28\\
SAM4MLLM & 8.52 & 2.55 & 5.65 & 4.57 & 24.74 & 9.20\\
PixelLM & 10.33 & 2.52 & 6.06 & 4.62 & 14.14 & 7.53\\
UniPixel & 13.90 & 2.89 & \underline{7.18} & 33.68 & 35.52 & 18.63\\
\midrule
\rowcolor{ourswash}
\textbf{MedUP} & \underline{24.59} & \underline{18.76} & 4.67 & \textbf{76.29} & \textbf{83.36} & \textbf{41.54}\\
\bottomrule
\end{tabular*}
\endgroup
\end{table}

\paragraph{Offline prefix allocation at a matched budget.}
Tab.~\ref{tab:allocation_policy} compares prefix allocation policies across 2,791 reference regions and 4,056 downstream segmentation queries from ACDC, AMOS2022-MR, and BTCV. Increasing a uniform fixed training budget from $K=2$ to $K=8$ improves oracle tokenizer reconstruction from 86.82 to 90.83 Dice, yet degrades downstream VLM segmentation from 34.00 to 30.26 Dice because forcing eight codes onto simple masks increases autoregressive sequence error. Conversely, shuffling our adaptive lengths across masks preserves the average budget ($\bar K = 4.612$) but causes 37.76\% of masks to violate the fidelity tolerance and drops reconstruction Dice from 91.19 to 87.43. Fidelity-constrained selection resolves both issues simultaneously, achieving zero fidelity violations, the highest reference reconstruction (91.19 Dice), and a 9.82-point downstream segmentation gain over length-matched shuffling (40.30 vs.\ 30.48 Dice).

\subsection{Mutual Gains in Understanding \& Perception}
\label{sec:exp_supervision}

\definecolor{abSeg}{HTML}{DFEBF3}
\definecolor{abQA}{HTML}{E9E2F1}
\definecolor{abSup}{HTML}{EDF0F3}
\begin{table}[t]
\centering
\vspace{-1mm}
\caption{\textbf{Training-supervision ablation across separately trained checkpoints.} $\mathrm{U}{\to}\mathrm{S}$ adds a target description before segmentation. $\mathrm{S}{\to}\mathrm{U}$ adds segmented evidence before the QA answer. Each $\Delta$ measures the change from enabling the indicated supervision with the other held fixed. Evaluation uses 10,073 segmentation queries and 5,844 QA items. Scores are percentages and changes are percentage points.}
\label{tab:factorial}
\begingroup
\footnotesize
\setlength{\tabcolsep}{0pt}
\renewcommand{\arraystretch}{1}
\begin{tabular}{
    >{\centering\arraybackslash}p{0.10\linewidth}
    >{\centering\arraybackslash}p{0.10\linewidth}
    >{\centering\arraybackslash}p{0.19\linewidth}
    >{\centering\arraybackslash}p{0.19\linewidth}
    >{\centering\arraybackslash}p{0.19\linewidth}
    >{\centering\arraybackslash}p{0.19\linewidth}
}
\toprule
\multicolumn{2}{c}{\cellcolor{abSup}\textbf{Supervision}} &
\multicolumn{2}{c}{\cellcolor{abSeg}\textbf{Segmentation}} &
\multicolumn{2}{c}{\cellcolor{abQA}\textbf{Understanding}}\\
\cmidrule(lr){1-2}
\cmidrule(lr){3-4}
\cmidrule(lr){5-6}
$\mathrm{U}{\to}\mathrm{S}$ & $\mathrm{S}{\to}\mathrm{U}$
  & Dice $\uparrow$ & $\Delta_{\mathrm{U}\to\mathrm{S}}$
  & Acc. $\uparrow$ & $\Delta_{\mathrm{S}\to\mathrm{U}}$\\
\midrule
--  & --  & 62.45 & --   & 62.80 & --\\
Yes & --  & 63.82 & +1.37 & 63.04 & --\\
--  & Yes & 62.91 & --   & 64.24 & +1.44\\
\rowcolor{ourswash}
Yes & Yes & 64.36 & +1.45 & 64.92 & +1.88\\
\bottomrule
\end{tabular}
\par\vspace{3pt}
\endgroup
\end{table}
\vspace{-1mm}
\begin{figure}[t]
    \centering
    \includegraphics[width=0.95\columnwidth]{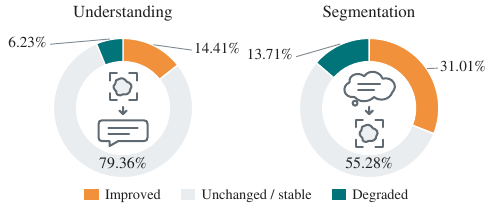}
    \vspace{-4mm}
    \caption{\textbf{How evidence and target descriptions change predictions.} Left, evidence before answering corrects 842 wrong answers while overturning 364 correct ones, a net 8.18-point gain over 5,844 questions. Right, target descriptions raise Dice by over one point on 31.01\% of 10,073 queries and lower it on 13.71\%.}
    \label{fig:reasoning_changes}
\end{figure}

\paragraph{Factorial ablation of training supervision.}
We train four checkpoints separately on the same paired examples, with the same background tasks and number of updates (Tab.~\ref{tab:factorial}, Supp.~\textcolor{cvprblue}{B}). Adding target descriptions to segmentation supervision raises Dice by 1.37 points with text-only QA supervision and by 1.45 points with region-evidence QA supervision. Training on QA responses with region spans raises answer accuracy by 1.44 percentage points without target descriptions and by 1.88 points with them. The checkpoint trained with both forms of supervision reaches 64.36 Dice and 64.92\% QA accuracy.
\vspace{-4mm}

\paragraph{Inference-time thinking vs.\ direct prediction.}
At the same final checkpoint, we compare direct prediction with describing the target before segmentation or generating evidence before answering (Fig.~\ref{fig:reasoning_changes}). Across 10,073 queries, segmentation Dice rose by more than one point on 31.01\%, fell by more than one point on 13.71\%, and stayed within one point on 55.28\%. Across 5,844 QA items, 842 incorrect answers became correct and 364 correct answers became incorrect, a net gain of 8.18 percentage points (Fig.~\ref{fig:reasoning_changes}).

\subsection{External Generalization on MeCoVQA}
\label{sec:exp_ood}
\vspace{-1mm}

We evaluate zero-shot segmentation on 1,243 MeCoVQA queries after excluding identifiable case overlap with training data and MedGround-Bench (Supp.~\textcolor{cvprblue}{D}). MedUP reaches 41.54 Dice and 35.47 IoU averaged equally across five modalities, 4.17 and 3.64 points above UniBiomed (Tab.~\ref{tab:mecovqa_ood} and Supp.~Tab.~\textcolor{cvprblue}{S8}). MedUP obtains 76.29 Dice on fundus photography and 83.36 on endoscopy, respectively 8.43 and 17.27 points above UniBiomed. MedUP ranks second on MRI and X-ray, but scores 4.67 Dice on ultrasound nerve segmentation, compared with 33.94 for MedPLIB. All 169 ultrasound queries target the brachial plexus, whereas none of the eight ultrasound sources in tokenizer training lists peripheral nerves as a target (Supp.~Tabs.~\textcolor{cvprblue}{S6} and~\textcolor{cvprblue}{S7}). This anatomical shift helps explain the low Dice on this subset.

\section{Conclusion and Future Work}
\label{sec:conclusion}
% MedUP generates an average of 3.32 residual codes per region, 58.5\% fewer than the fixed budget. In paired training ablations, adding target descriptions improves segmentation Dice, while adding region evidence to QA supervision improves answer accuracy. The current tokenizer supports up to eight codes per mask. Extending this limit would let us test whether additional codes improve boundary reconstruction for small or irregular structures. Future work could also test whether the gains observed with Qwen3-VL-4B hold across other backbones, model sizes, and training regimes. Our evaluation uses offline benchmarks, including held-out MeCoVQA cases. Prospective studies could assess how well MedUP incorporates clinicians' corrections, tracks findings across follow-up scans, and handles changes in acquisition protocols.

% MedUP averages 3.32 residual codes per region, 58.5\% fewer than the fixed eight-code budget. In paired training ablations, target descriptions improve segmentation Dice, and region-evidence supervision improves QA accuracy. Future work could test whether raising the tokenizer's eight-code limit improves boundary reconstruction for small or irregular structures and whether the gains with Qwen3-VL-4B hold across backbones, model sizes, and training regimes. We evaluated MedUP on offline benchmarks, including held-out MeCoVQA cases. Prospective studies could assess how well MedUP incorporates clinicians' corrections, tracks the same finding across follow-up scans, and maintains accuracy when acquisition~protocols~change.

MedUP bridges clinical reasoning and anatomical perception through fidelity-constrained adaptive region tokens, averaging 3.32 residual codes per region (58.5\% below the eight-code budget) while confirming that target descriptions improve segmentation Dice and region evidence boosts QA accuracy. Future work will explore larger code budgets for fine structures and broader VLM backbones, alongside prospective clinical studies on interactive clinician corrections, longitudinal lesion tracking, and robustness to cross-site acquisition and annotation shifts.

\section*{LLM Usage Statement}
We used a large language model to edit the English prose. During dataset construction, we used vision--language models to generate annotated-region descriptions and propose candidate evidence boxes for OmniMedVQA questions (Supp.~\textcolor{cvprblue}{E} and~\textcolor{cvprblue}{F}). We designed the method, conducted the experiments, and interpreted the results. We take responsibility for the manuscript and its scientific conclusions.

% END merged from sec/06_conclusion.tex
% END merged from body.tex
\end{document}